\documentclass[]{cit_lab_mfr}

\usepackage{amssymb}

\usepackage{amsthm}
\usepackage{caption}
\usepackage{amsmath}

\usepackage{mathtools}

\usepackage{enumitem}

\usepackage{hyperref}
\usepackage{cleveref}
\usepackage{verbatim}

\usepackage{wrapfig}  
\usepackage{graphicx}
\usepackage{floatrow}
\usepackage{subcaption}
\usepackage{listings}
\usepackage{algorithm}

\usepackage{microtype}
\usepackage{graphicx}
\usepackage{subcaption}
\usepackage{booktabs} 
\usepackage{multirow}
\usepackage{graphicx}
\usepackage[normalem]{ulem}
\usepackage{enumitem}
\useunder{\uline}{\ul}{}
\usepackage[utf8]{inputenc}
\usepackage{xcolor}
\usepackage{graphicx}
\providecommand{\XSolidBrush}{\ensuremath{\times}}
\providecommand{\Checkmark}{\ensuremath{\checkmark}}
\definecolor{DarkGreen}{HTML}{006400}
\definecolor{DarkRed}{HTML}{8B0000}
\definecolor{diy_pink}{RGB}{255,247,240}

\usepackage{multirow}         
\usepackage[table]{xcolor}     

\usepackage{graphicx}   
\usepackage{adjustbox}  
\usepackage{array}      
\usepackage{graphicx}
\usepackage{caption} 
\usepackage{xcolor}
\usepackage{pifont}

\usepackage{hyperref}

\usepackage{subcaption} %

\usepackage[toc,page,header]{appendix}

\usepackage{minitoc}

\usepackage{amsmath}

\usepackage{mathtools}

\usepackage{xurl}

\RequirePackage{algorithmic}

\title{Fysiverse-3D-Vision Technical Report: Generating Executable \\ 3D Worlds from Images through Unified Spatial Reasoning}

\author{%
\parbox{\textwidth}{\centering
Dingkang Yang$^{\dagger}$, Yizhou Liu, Wendong Cheng, Zizhi Chen, Shunli Wang, \\Yang Liu, Hongsheng Li$^{\S}$, Lihua Zhang$^{\S}$

}}

\affiliation{%
\parbox{\textwidth}{\centering\small
Physical Superintelligence Lab, Fysics AI \\[1mm]
College of Intelligent Robotics and Advanced Manufacturing, Fudan University\\[1mm]
Multimedia Laboratory (MMLab), The Chinese University of Hong Kong\\[1mm]
College of Electronic and Information Engineering, Tongji University\\[1mm]
}}

\contribution[\dagger]{Project lead}
\contribution[\S]{Corresponding author}

\abstract{
Generative models have substantially advanced image-conditioned 3D content creation, yet generating controllable and executable 3D scenes from a single image remains challenging. Existing 3D generative approaches can synthesize visually plausible objects and scenes, but their spatial layout estimation is often tightly coupled with specific asset generators. Consequently, they struggle to jointly model object semantics, metric geometry, and scene-level spatial relationships, which are essential for interactive editing, physical simulation, and embodied applications. We propose Fysiverse-3D-Vision, a unified vision-language-geometry framework for generative 3D scene reconstruction and executable asset construction from a single image. The core idea is to establish a shared representation where spatial reasoning and geometric reconstruction mutually enhance each other, allowing object layouts to be inferred beyond the constraints of individual asset generators. Specifically, our model integrates textual supervision, semantic visual cues, and geometric representations within a unified Transformer to capture scene context, metric geometry, and object-level interactions. An object-conditioned layout module further performs cross-attention between target object representations and global geometric features to predict object translation, rotation, and scale. The training process progressively learns geometry-language alignment, introduces layout reasoning while preserving reconstruction capability, and refines physical consistency through collision-aware optimization. By separating spatial layout reasoning from asset synthesis, Fysiverse-3D-Vision provides an adaptable interface for interactive scene editing, object-level manipulation, and executable 3D content generation. Extensive experiments demonstrate that our framework achieves superior geometric consistency, layout estimation, rendering quality, and physical property understanding compared with existing approaches.

}
\date{\today}
\checkdata[Corresponding]{ \url{dicken@fyscis.ai}, \url{hsli@ee.cuhk.edu.hk}, \url{lihuazhang@fudan.edu.cn}}
\checkdata[Github]{\url{https://github.com/Fysics-AI/Fysiverse-3D-Vision}}
\checkdata[Hugging Face]{\url{https://huggingface.co/Fysics-AI/Fysiverse-3D-Vision}}

\begin{document}
\maketitle

\section{Introduction}

Recent advances in generative modeling are reshaping computer vision from recognition centered perception toward the construction of structured, controllable, and interactive 3D worlds. Beyond synthesizing visually realistic content, generative models are increasingly expected to recover latent scene structures and produce representations that support editing, simulation, and embodied interaction. In embodied intelligence, digital simulation, and immersive applications~\cite{zhou2020distractor,wang2024restoring,yoon2026splitflow,rao20253dpr,yuan2025scaling,zhou2026stream3d}, a central challenge is to transform visual observations into executable 3D scenes, where objects are jointly represented through appearance, metric geometry, spatial relationships, and physical functionality. However, conventional recognition and reconstruction paradigms often focus on isolated semantic prediction or geometric recovery~\cite{liu2026resolving,han2026omnifysics,Xue_2026_CVPR,chen2026forging}, making it difficult to generate coherent scene structures and support subsequent interaction under incomplete visual observations.

Generating 3D scenes from a single image~\cite{midi,scenegen,3d-front,3d-future} represents an important step toward this goal. A successful system needs to recover not only the appearance and geometry of individual objects, but also their spatial organization within a shared environment~\cite{i-scene,sam3d}. This problem is particularly challenging because a single viewpoint provides limited information about object structure, depth, and physical relationships. Therefore, the model must jointly reason about object identity, metric geometry~\cite{vggt,pi3}, and inter-object configurations~\cite{layoutgpt,layoutvlm}, while handling occlusions, ambiguous observations, and diverse real-world appearances.
Existing approaches for image-based 3D scene generation can be broadly categorized into reconstruction-based methods~\cite{depr,gen3dsr} and retrieval-based methods~\cite{scenemaker,holodeck}. Reconstruction-based approaches typically leverage scene-level annotations~\cite{3d-front,3d-future} to estimate geometry, layouts, and object attributes directly, whereas retrieval-based approaches identify suitable assets from large databases~\cite{openshape,objaverse} and optimize their alignment with observed scenes. Although these methods have significantly advanced scene understanding, they remain limited by insufficient scene-level supervision, restricted asset diversity, or complicated multi-stage optimization procedures. Recent object-centric foundation models~\cite{trellis} provide a new direction by introducing strong 3D priors. Methods built upon these models~\cite{midi,scenegen,i-scene,sam3d} can extend object synthesis toward scene generation through additional interaction modeling and post-training. However, their layout reasoning capability is usually coupled with individual object generators, making spatial layouts dependent on specific synthesis models rather than serving as a general scene-level representation.

Beyond geometry generation, recent studies have explored extending object foundation models toward executable 3D assets by incorporating physical and functional knowledge. These efforts enable capabilities including physical property prediction~\cite{Physx-3d,PhysX-Anything}, part-level decomposition~\cite{Partcrafter,omnipart}, and articulated structure modeling~\cite{dreamart,SINGAPO}. Nevertheless, these capabilities are commonly introduced through task-specific post-training strategies. The resulting models often lack a unified representation that can simultaneously capture semantic understanding, spatial reasoning, geometric reconstruction, and executable attributes. Consequently, current methods still face difficulties when transferring from visual reconstruction to interactive scene generation, where object placement, physical properties, and functional structures need to be jointly considered. A more general framework is required to establish an intermediate spatial representation between scene perception and executable asset construction.
Recent advances in unified multimodal models~\cite{g2vlm} provide an opportunity to address this limitation. Unlike isolated feed-forward geometry predictors~\cite{vggt,moge}, unified models can integrate semantic reasoning and geometric perception within a shared representation space. Since spatial layouts depend on both object relationships and object-specific characteristics~\cite{Spatial-mllm}, language-guided reasoning offers complementary semantic knowledge beyond the implicit priors learned from large-scale object datasets~\cite{trellis,abo,Objaverse-xl,objaverse}. Therefore, a native vision-language-geometry model equipped with spatial understanding~\cite{qwen2} and geometric reconstruction ability~\cite{pi3} provides a promising foundation for building interactive and executable 3D environments.

\begin{figure}[t]
  \centering
  \includegraphics[width=\textwidth]{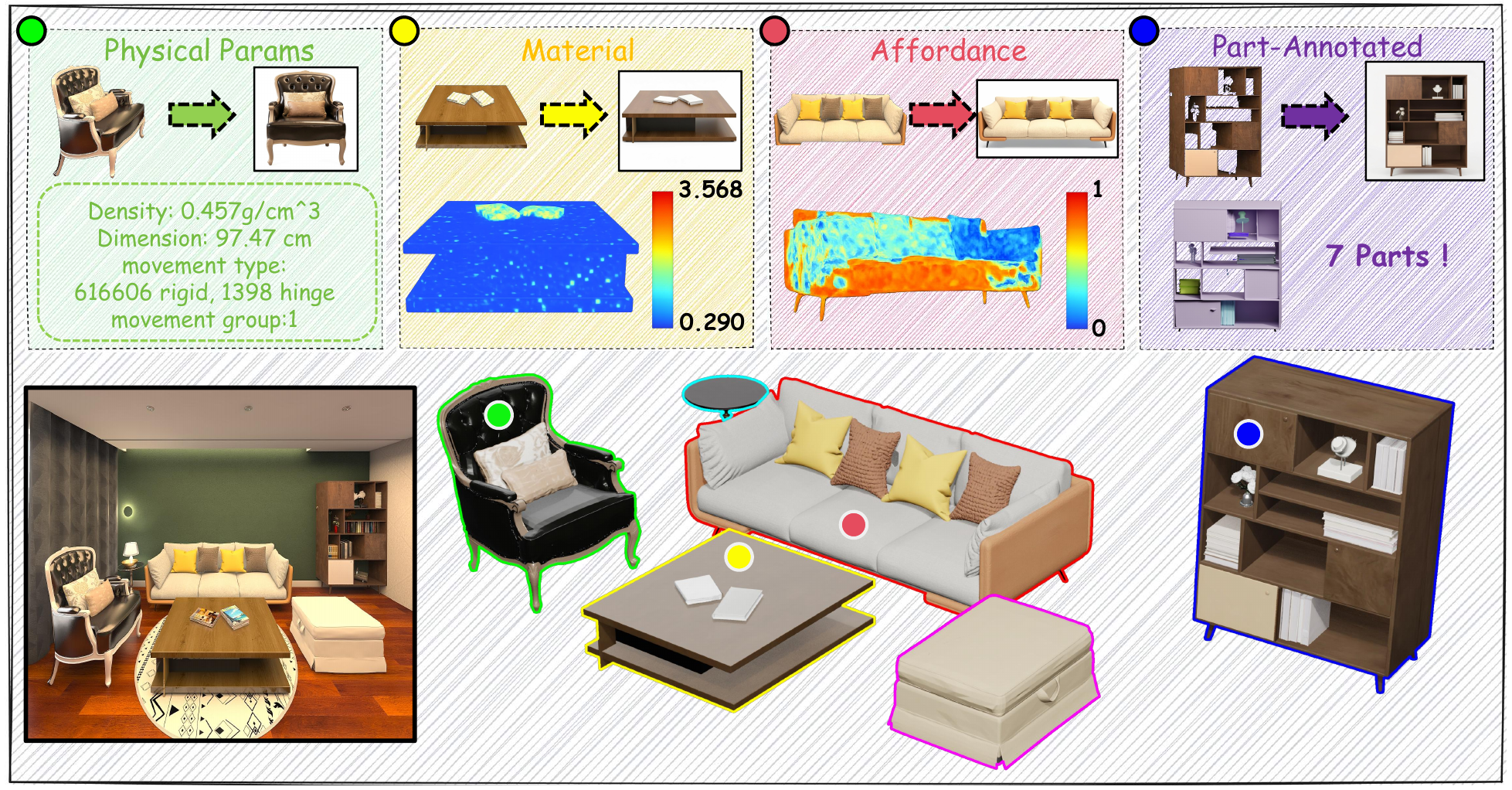}
  \caption{From a single image, Fysiverse-3D-Vision reconstructs individual objects, independently predicts their translation, rotation, and scale, and attaches physical, material, affordance, and part-level attributes for editing and simulation.}
  \label{fig1}
\end{figure}

Based on this insight, we propose Fysiverse-3D-Vision (F3V), a unified vision-language-geometry framework that enables executable 3D scene reconstruction by integrating spatial reasoning and geometric modeling. Instead of introducing an independent layout predictor trained only with scene annotations, our approach allows object placement to emerge from a shared representation learned through multimodal understanding and geometric reconstruction. We construct large-scale interleaved training samples through a simulation-based data generation pipeline and jointly encode textual tokens, semantic visual tokens, and geometric tokens within a single Transformer. This unified representation enables the model to preserve high-level scene semantics while recovering accurate 3D structures. Based on the learned spatial features, an object-conditioned layout module further establishes interactions between target object tokens extracted from scene observations and global geometry tokens, allowing direct prediction of object translation, rotation, and scale. Finally, we refine the predicted layouts with scene-level supervision and a collision-aware objective to improve physical validity. 

As shown in Figure~\ref{fig1}, our model decouples spatial layout reasoning from specific asset generators and establishes a flexible interface between scene understanding and executable asset construction. This design enables a wide range of downstream applications, including interactive scene editing, object manipulation, and simulation. By reasoning over shared semantic and geometric representations, F3V alleviates common challenges in single-image reconstruction, such as incomplete object observations, reconstruction artifacts, and inconsistent spatial configurations.

\section{Related Work}

\subsection{3D Scene Generation}
Image-based 3D scene generation aims to recover structured and interactive environments from visual observations, serving as an important foundation for embodied intelligence, simulation, and immersive content creation. Existing approaches mainly differ in how 3D objects are obtained and organized. Retrieval-based methods construct scenes by searching suitable assets from large-scale 3D repositories~\cite{3d-front,3d-future} and leveraging vision-language models (VLMs)~\cite{layoutgpt,layoutvlm} to infer semantic layouts and object relationships. Although these methods benefit from high-quality existing assets, their performance is inherently limited by the coverage, diversity, and availability of the underlying databases, which restricts their adaptation to open-world scenarios.
Recent generation-based approaches attempt to directly synthesize 3D content from images without relying on predefined asset collections. CAST~\cite{cast} introduces component-aware scene reconstruction with SDF-based physical refinement to improve structural consistency. MIDI~\cite{midi} and SceneGen~\cite{scenegen} explore single-pass generation of multiple objects while modeling implicit interactions and instance-level poses. PartCrafter~\cite{Partcrafter} further extends compositional 3D generation by jointly modeling object and part structures through latent diffusion transformers. More recent methods, including I-Scene~\cite{i-scene}, 3D-Fixer~\cite{3D-Fixer}, and SAM3D~\cite{sam3d}, focus on improving scalability and generalization through large-scale training data and diversified scene synthesis strategies. Despite these advances, existing approaches typically regard spatial reasoning and geometric generation as separate components. The lack of a unified representation that jointly captures scene semantics, geometry, and object interactions remains a major challenge for executable and controllable 3D scene understanding.

\subsection{Executable Asset Generation}
The increasing demand for embodied agents and physical simulation has shifted 3D generation from visual realism toward executable assets with explicit structures, physical properties, and interaction capabilities. Recent studies have explored enriching generated objects with functional and physical information. PartPacker~\cite{PartPacker} improves part-level generation efficiency through a dual-voxel diffusion architecture, enabling automatic part discovery without requiring explicit segmentation. OmniPart~\cite{omnipart} further introduces part-aware latent modeling under bounding-box constraints, supporting structured generation with improved editability and spatial control. For articulated objects, DreamArt~\cite{dreamart} utilizes generated videos to optimize movable object structures, while URDF-Anything~\cite{URDF-Anything} directly predicts URDF representations for robotic simulation. However, these approaches still depend on specific annotations or input conditions, and often lack comprehensive modeling of appearance, physical attributes, and interaction-related properties.
Physics-aware asset generation methods have recently attempted to bridge the gap between visual reconstruction and simulation requirements. Existing solutions incorporate material characteristics and physical parameters, but they usually treat different object categories independently or focus on limited physical factors. PhysXGen~\cite{Physx-3d} introduces a unified framework for generating 3D assets with physical attributes, including size and density. Based on this direction, PhysX-Anything~\cite{PhysX-Anything} extends physical asset generation to real-image inputs and produces simulation-ready objects with explicit physical properties. Nevertheless, executable asset generation still requires accurate scene-level understanding, since object functionality depends not only on individual asset properties but also on their spatial context and relationships within the environment.

\subsection{Unified 3D Modeling}

Recent advances in unified multimodal foundation models~\cite{Qwen35-omni,Longcat-flash-omni,Ming-omni,Show-o2,unigen,T2i-r1} have demonstrated a transition from isolated task-specific architectures toward shared understanding and generation paradigms. BLIP3-o~\cite{Blip3-o} shows that separating multimodal understanding and generation through an ``understand first, generate later'' strategy can improve cross-modal alignment, while Bagel~\cite{begal} introduces a Mixture-of-Transformer-Experts (MoT) design to reduce interference among heterogeneous objectives.
Inspired by these developments, unified modeling has also emerged in the 3D domain. ShapeLLM-Omni~\cite{Shapellm-omni} and Omni123~\cite{omni123} investigate native 3D foundation models by introducing discrete 3D representations and jointly training 2D and 3D modalities. Omni-View~\cite{Omni-view} combines geometry and texture representations to unify scene understanding, novel-view synthesis, and geometric estimation. UniUGG~\cite{Uniugg} and G2VLM~\cite{g2vlm} further incorporate geometric information into multimodal language models, enabling stronger spatial reasoning and reconstruction capabilities. However, current unified 3D models mainly focus on representation learning or object-level generation, while the potential of shared understanding-generation architectures for scene-level spatial reasoning and executable layout prediction remains largely unexplored.

\begin{figure}[t]
 \centering
 \includegraphics[width=\linewidth]{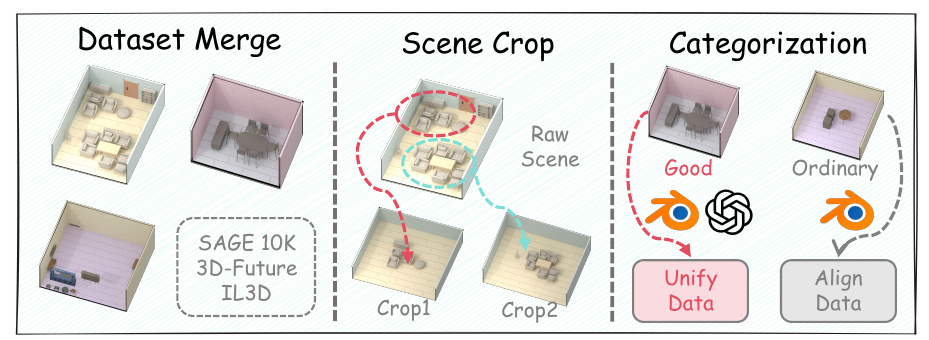}
\caption{Three-stage simulation-data curation pipeline: heterogeneous indoor datasets are merged, compact object-centric regions are cropped, and samples are categorized by structural completeness, visibility, and semantic quality.}
\label{fig3}
\vspace{-5pt}
\end{figure}

\section{Methodology}
\subsection{Data Governance Procedure}
As illustrated in Figure~\ref{fig3}, we develop a three-stage data preparation pipeline that transforms heterogeneous 3D resources into structured training samples. The pipeline consists of unified multi-source data integration, object-centric region extraction, and quality-aware sample refinement. Specifically, we collect indoor scene data from 3D-FUTURE~\cite{3d-future}, SAGE-10K~\cite{sage}, and IL3D~\cite{il3d}, covering a broad range of room layouts, furniture configurations, and spatial arrangements. After applying consistent preprocessing, the resulting dataset contains approximately 46K spatial instances, 189K object instances, and 254K multi-view rendered images, providing diverse supervision for spatial understanding and geometric reasoning.

Directly using raw indoor scenes is challenging due to their large spatial extent, uneven object distributions, and limited informative regions. To obtain compact regions suitable for model learning, we construct a spatial relation graph where object centers are treated as nodes and object pairs are connected according to their 3D distances~\cite{graph}. We first identify spatially coherent object groups through graph connected-component analysis and then apply DBSCAN clustering~\cite{DBSCAN} to locate dense regions within these groups. Candidate regions are selected according to object quantity, spatial compactness, density, and overlap constraints. The retained regions are subsequently normalized by re-centering objects, aligning scene scales, completing consistent room boundaries, and rendering additional views with optimized camera configurations. This process improves object visibility and provides more informative observations for spatial reasoning.
After region extraction, we further perform quality-based filtering according to structural completeness, object arrangement, rendering quality, and semantic relevance. Samples that satisfy all criteria are directly preserved, while partially qualified samples are geometrically aligned and normalized before inclusion. The resulting dataset maintains the diversity of multi-source 3D environments while providing clean and structured scene representations for unified spatial understanding.

\begin{figure}[t]
 \centering
 \includegraphics[width=\linewidth]{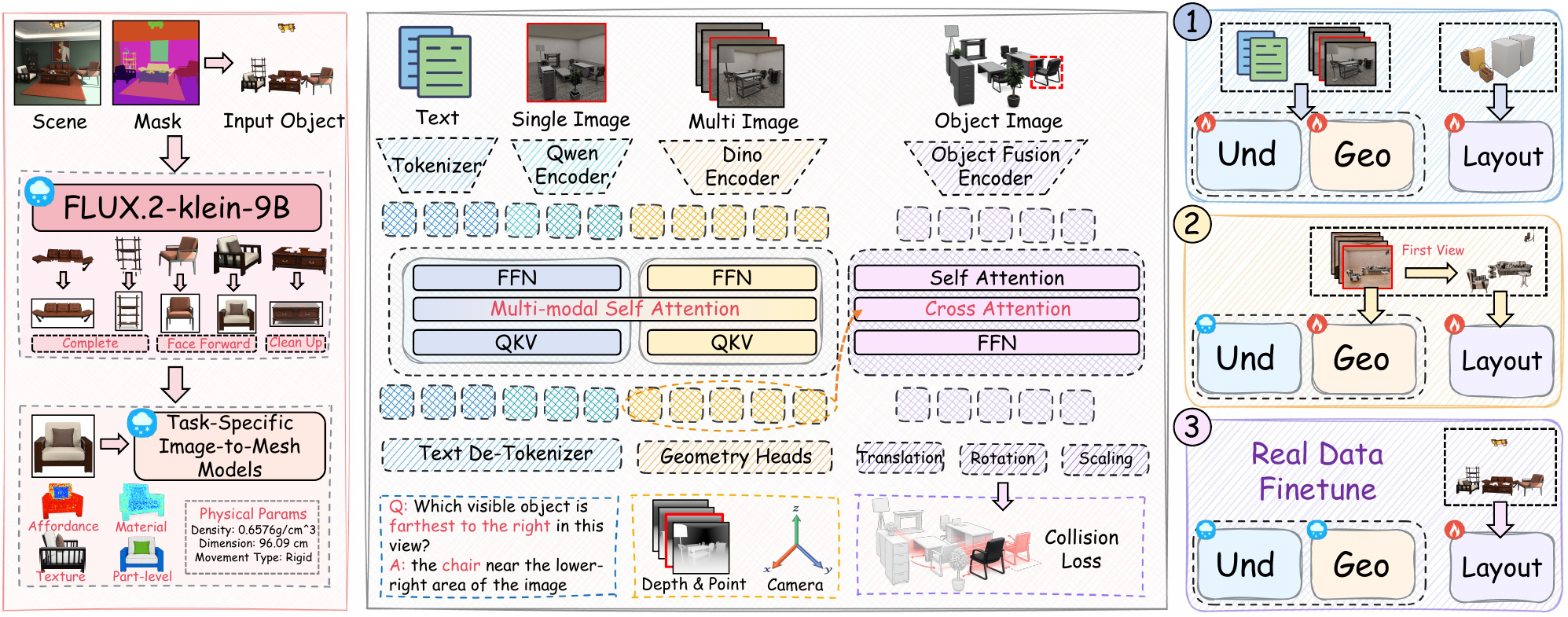}
\caption{Our method decouples executable asset generation from 3D layout prediction. 
Completed object crops are used to generate textured, physical, and part-level assets, while a unified vision-language-geometry Transformer predicts object translation, rotation, and scale from scene geometry and object-condition tokens. 
Training proceeds through geometry-language pretraining, layout injection, and real-data refinement.}
\label{fig2}
\vspace{-5pt} 
\end{figure}

\subsection{Unified Geometry-Language Representation}
As shown in Figure~\ref{fig2}, F3V builds upon a unified 3D vision-language model, where semantic understanding and geometric perception are learned within the same representation space. Object layout estimation requires more than predicting object coordinates from masks. A reliable layout should simultaneously capture object identity, scene context, and metric spatial structure. Learning these capabilities independently often leads to insufficient interaction between semantic reasoning and geometric reconstruction. Therefore, we adopt a unified backbone that enables text tokens, semantic visual tokens, and geometric visual tokens to interact through shared multimodal self-attention.
Given text tokens $\mathbf{X}^{txt}$, semantic visual tokens $\mathbf{V}^{sem}$, and geometric visual tokens $\mathbf{V}^{geo}$, the unified input sequence is
$
\mathbf{X}_0=[\mathbf{X}^{txt};\mathbf{V}^{sem};\mathbf{V}^{geo}]
$.
All modalities are fused via unified \textbf{M}ulti-\textbf{M}odal \textbf{S}elf-\textbf{A}ttention:
\begin{equation}
\mathbf{X}_{l+1}
=
\operatorname{MMSA}(\mathbf{X}_{l}),
\qquad
\mathbf{H}=\mathbf{X}_{Last}.
\end{equation}
The shared attention allows different modalities to contribute complementary information. Specifically, semantic tokens capture object categories, language instructions, and scene-level context, while geometric tokens encode depth, camera information, and spatial structures. Their joint interaction produces hidden representations that preserve both semantic awareness and geometric reasoning ability.
The geometric branch decodes $\mathbf{H}$ into local point maps, global point maps, and camera poses:
\begin{equation}
(\hat{\mathbf{P}}^{loc},\hat{\mathbf{P}}^{glob},\hat{\mathbf{G}})
=
\mathcal{D}_{geo}(\mathbf{H}).
\end{equation}
These geometric predictions provide explicit structural supervision rather than serving only as auxiliary outputs. By constraining the shared representation with 3D reconstruction objectives, the model acquires geometry-aware features that can be further utilized for downstream layout prediction.

\subsection{Object-Conditioned Layout Branch}
The layout prediction module is introduced on top of the geometry-aware hidden representations. Instead of estimating object placement directly from image-level features, the proposed design explicitly incorporates object-specific conditions and scene-level geometric information. For clarity, the reference RGB image is denoted as $R$. Given $R$, a binary target mask $M\in{0,1}^{H\times W}$, and a point map $P\in\mathbb{R}^{H\times W\times3}$ aligned with $R$~\cite{moge}, the object condition encoder $C$~\cite{sam3d} extracts object-centric representations and maps them into the unified hidden space:
\begin{equation}
(\tilde{\mathbf{O}},\mathbf{Q}) = \big(\Phi(\mathbf{O}),\mathbf{Q}\big), 
\quad (\mathbf{O},\mathbf{Q})=C(R,M,P,R\odot M),
\end{equation}
where $C$ is the condition encoder, $\mathbf{O}\in\mathbb{R}^{B\times K\times d_o}$ denotes the object tokens before projection, $\tilde{\mathbf{O}}\in\mathbb{R}^{B\times K\times d}$ denotes the aligned object tokens in the unified hidden space, $K$ is the number of object tokens, and $\mathbf{Q}$ gives the corresponding object-token positions.

The layout decoder operates on geometry representations associated with the reference view. Specifically, $\mathbf{G}_{1}$ denotes the first-view geometry tokens selected from the unified hidden states $\mathbf{H}$, while $\mathbf{U}_{1}$ represents their corresponding token positions. The decoder establishes object-to-scene interactions through cross-attention, where projected object tokens are used as queries and reference-view geometry tokens provide keys and values:
\begin{equation}
\mathbf{Z}=
\operatorname{Softmax}
\left(
\frac{(W_q\tilde{\mathbf{O}})(W_k\mathbf{G}_{1})^{\top}}{\sqrt{d}}
\right)
W_v\mathbf{G}_{1}.
\end{equation}
Here, $W_q$, $W_k$, and $W_v$ are learnable query, key, and value projections, respectively. $d$ denotes the attention dimension, while token positions $\mathbf{Q}$ and $\mathbf{U}_{1}$ provide positional information for layout decoding. Although the decoder uses geometry tokens from a single reference view, $\mathbf{G}_{1}$ does not represent an isolated observation. Before being selected, these tokens have already exchanged information with multi-view geometry features through the shared self-attention mechanism in the unified backbone.
The layout heads predict object translation, rotation, and scale from the fused representation:
\begin{equation}
\hat{\mathbf{t}}=f_t(\mathbf{Z}),\qquad
\hat{\mathbf{r}}=f_r(\mathbf{Z}),\qquad
\hat{\mathbf{s}}=f_s(\mathbf{Z}),
\end{equation}
where $\hat{\mathbf{t}}\in\mathbb{R}^{3}$ represents the object-center translation, $\hat{\mathbf{r}}\in\mathbb{R}^{3\times3}$ denotes the 9D raw rotation prediction, and $\hat{\mathbf{s}}\in\mathbb{R}^{+}$ indicates the object scale. We initialize the layout decoder from the camera decoder and initialize $f_t$ and $f_r$ from the corresponding camera pose heads. This transfers rigid transformation priors learned from camera estimation to object-level pose prediction.

\subsection{Multi-Stage Training Strategy}

\noindent \textbf{Stage 1: Learning a Shared Geometry-Language Space.}
The first training stage focuses on establishing a unified representation before introducing the layout prediction module. The objective is to construct a latent space that simultaneously captures semantic reasoning and metric 3D structures, providing a foundation for subsequent object placement prediction.
The understanding branch is optimized with image-text question answering supervision. Given the input frames $I$, question text $T$, and answer sequence $a$, the language decoder is trained through next-token prediction:
\begin{equation}
\mathcal{L}_{CE}
=
-\sum_{i=1}^{L} \log p_{\theta}(a_i|a_{<i},I,T).
\end{equation}
where $a_i$ denotes the $i$-th answer token, and $p_{\theta}$ represents the token distribution generated by the language decoder. This objective maintains the pretrained VLM capability in object recognition, instruction following, and spatial relation reasoning.

Meanwhile, the geometry branch receives multi-view RGB observations together with depth and camera supervision. The purpose is to encourage the shared latent space to encode metric scene structures using the $\pi^3$~\cite{pi3} objective:
\begin{equation}
\mathcal{L}_{VG}
=
\lambda_l\mathcal{L}_{loc}
+
\lambda_g\mathcal{L}_{glob}
+
\lambda_c\mathcal{L}_{cam},
\end{equation}
where $\mathcal{L}_{loc}$ and $\mathcal{L}_{glob}$ constrain local and global point-map reconstruction, respectively, and $\mathcal{L}_{cam}$ supervises cross-view camera pose estimation.
Through joint optimization of semantic and geometric objectives, Stage 1 learns a representation that combines category-level understanding with coordinate-aware spatial perception. The overall objective of this stage is formulated as
$
\mathcal{L}_{s1}
=
\mathcal{L}_{CE}
+
\mathcal{L}_{VG}
$.
After obtaining the shared geometry-language representation, we freeze the backbone and train the newly introduced layout module using alignment data~\cite{il3d}. This warm-up process aligns object-conditioned features with the geometry-aware latent space before full optimization. For an object with ground-truth layout $(\mathbf{t}^{*}, \mathbf{r}^{*}, \mathbf{s}^{*})$, the predicted layout $(\hat{\mathbf{t}}, \hat{\mathbf{r}}, \hat{\mathbf{s}})$ is supervised using the Smooth-L1~\cite{fast-rcnn} penalty $\zeta_{\delta}(\cdot)$:
\begin{equation}
\mathcal{L}_{t}
=
\zeta_{\delta}(\hat{\mathbf{t}}-\mathbf{t}^{*}),
\quad
\mathcal{L}_{r}
=
\left\langle
\mathbf{W}_{r},
\zeta_{\delta}(\hat{\mathbf{r}}-\mathbf{r}^{*})
\right\rangle,
\quad
\mathcal{L}_{s}
=
\zeta_{\delta}(\hat{\mathbf{s}}-\mathbf{s}^{*}),
\end{equation}
Translation and scale are optimized through direct regression, while rotation prediction adopts an element-wise weighting matrix $\mathbf{W}_{r}$ to emphasize yaw-related components under the $z$-up assumption. The operator $\langle \cdot, \cdot \rangle$ represents the averaged weighted element-wise summation. The alignment objective is defined as:
\begin{equation}
\mathcal{L}_{\mathrm{align}}
=
\lambda_t \mathcal{L}_{t}
+
\lambda_r \mathcal{L}_{r}
+
\lambda_s \mathcal{L}_{s},
\end{equation}
where $\lambda_t$, $\lambda_r$, and $\lambda_s$ balance the contributions of translation, rotation, and scale terms.

\noindent \textbf{Stage 2: Layout Injection with Geometry Preservation.}
After establishing the shared geometry-language representation, we introduce the layout branch while maintaining the geometry reconstruction objective~\cite{sage,3d-future}. The overall optimization target is formulated as
$
\mathcal{L}_{s2}
=
\mathcal{L}_{VG}
+
\lambda_{a}\mathcal{L}_{align}
$.
During this stage, the layout module infers object translation, rotation, and scale from the interaction between $\mathbf{G}_{1}$ and $\tilde{\mathbf{O}}$ under the supervision of $\mathcal{L}_{a}$. Meanwhile, retaining the geometry reconstruction loss $\mathcal{L}_{VG}$ prevents the shared representation from degrading toward layout-specific optimization and maintains the original 3D reconstruction capability.

\noindent \textbf{Stage 3: Collision-Aware Layout Refinement.}
In the final stage, we keep the unified backbone fixed and optimize only the layout-related parameters. Since the geometry-aware representation has already been sufficiently established, this stage focuses on improving the physical validity and spatial plausibility of object placement. The layout regression objective remains consistent with $\mathcal{L}_{align}$.
To further reduce physically implausible object intersections, we introduce a BEV-based collision constraint:
\begin{equation}
\mathcal{A}^{bev}_{j}
=
\left|
\Pi_{bev}(\hat{\mathcal{B}})
\cap
\Pi_{bev}(\mathcal{B}_j)
\right|,
\end{equation}
\begin{equation}
\mathcal{L}_{col}
=
\sum_j
\mathbb{I}[\Delta_z(\hat{\mathcal{B}},\mathcal{B}_j)>0]\,
\mathcal{A}^{bev}_{j},
\end{equation}
where $\hat{\mathcal{B}}$ is the predicted oriented bounding box of the target object, $\mathcal{B}_j$ represents the bounding box of the $j$-th neighboring object, and $\Pi_{bev}(\cdot)$ projects a 3D bounding box onto the bird's-eye-view plane. $\mathcal{A}^{bev}_{j}$ measures the intersection area between the two projected footprints. The term $\Delta_z(\hat{\mathcal{B}},\mathcal{B}_j)$ evaluates vertical overlap, ensuring that the collision penalty is applied only when two objects intersect along the height direction.
The final optimization objective is defined as:
\begin{equation}
\mathcal{L}_{s3}
=
\mathcal{L}_{align}
+
\lambda_{col}\mathcal{L}_{col},
\end{equation}
where $\lambda_{col}$ controls the contribution of the collision constraint.

\subsection{Inference Pipeline}
During inference, F3V takes a scene image and a target object mask as input. If the mask is unavailable, an open-vocabulary segmentation model is first applied to identify object regions~\cite{ground-sam2}. The extracted mask is decomposed into individual object components, and each component is used to generate a masked object crop. To obtain a complete 3D asset, the masked crop is transformed into a clean frontal representation, followed by textured mesh reconstruction $\mathcal{M}$ using existing 3D generation methods~\cite{trellis2,hunyuan3d,Physx-3d,Partcrafter}.
The reconstructed mesh is responsible only for recovering object appearance and geometry, whereas its spatial configuration is determined by the proposed layout model based on the scene image and object mask. We construct an intermediate scene representation containing the reference image, target mask, object category, and reconstructed mesh information. The trained model then predicts the object transformation parameters:
\begin{equation}
(\hat{\mathbf{t}},\hat{\mathbf{r}},\hat{\mathbf{s}})
=
F_{\theta}(R,M),
\end{equation}
Finally, the generated object instance is obtained by applying the predicted similarity transformation to each vertex $\mathbf{v}$ of the reconstructed mesh $\mathcal{M}$:
\begin{equation}
\hat{\mathcal{M}}
=
\{\hat{\mathbf{s}}\hat{\mathbf{r}}\mathbf{v}+\hat{\mathbf{t}}
\mid
\mathbf{v}\in\mathcal{M}\}.
\end{equation}

\section{Experiments}

\subsection{Experimental Setting}

\noindent \textbf{Evaluation Metrics.}
We assess the reconstructed scenes from both geometric accuracy and visual fidelity. For geometric evaluation, point clouds are sampled from the generated asset surfaces, and the predicted scenes are registered to the ground truth using FilterReg. Compared with ICP, FilterReg provides stronger robustness when handling incomplete observations and inaccurate initial poses, which are common in multi-object reconstruction scenarios. After registration, we measure scene-level and object-level reconstruction performance using Chamfer Distance and F-Score, denoted as CD-S, F-Score-S, CD-O, and F-Score-O, respectively. We additionally report the voxel IoU of object bounding boxes (IoU-B) to evaluate spatial occupancy consistency and layout accuracy. For visual assessment, the aligned predictions are rendered from the original input viewpoint in Blender and compared with reference images using PSNR, SSIM, LPIPS, and CLIP-S. These metrics evaluate pixel-level similarity, structural preservation, perceptual quality, and semantic alignment. We also record the inference latency required to generate a single 3D asset on one H200 GPU to analyze computational efficiency.

\begin{table*}[t]
\caption{
\label{tab1}
Quantitative evaluation of image-based 3D scene reconstruction on the 3D-FUTURE test set.
The comparison covers geometric accuracy, spatial layout consistency, and visual reconstruction quality. $^*$ indicates adopting MV-Adapter~\cite{Mv-adapter} for texture rendering.
}
\centering
\footnotesize
\centering
\setlength{\tabcolsep}{2pt}
\resizebox{1\textwidth}{!}{%
\begin{tabular}{@{}lc|ccccc|cccc@{}}
\toprule
\multirow{2}{*}{\bf Methods} &
\multirow{2}{*}{\bf Inst. Spec.} &
\multicolumn{5}{c|}{\bf Geometry} &
\multicolumn{4}{c}{\bf Visual Metrics} \\
& &
{\bf CD-S $\downarrow$} &
{\bf CD-O $\downarrow$} &
{\bf F-Score-S $\uparrow$} &
{\bf F-Score-O $\uparrow$} &
{\bf IoU-B $\uparrow$} &
{\bf PSNR $\uparrow$} &
{\bf SSIM $\uparrow$} &
{\bf LPIPS $\downarrow$} &
{\bf CLIP-S $\uparrow$} \\
\midrule
PartCrafter~\cite{Partcrafter}
& \XSolidBrush
& 0.2027 & --- & 35.86 & --- & ---
& --- & --- & --- & ---   \\

Gen3DSR~\cite{gen3dsr}
& \Checkmark
& 0.1260 & 0.1511 & 43.81 & 38.95 & 0.3162
& 10.15 & 0.7874 & 0.4225  & 0.5972  \\

MIDI$^*$~\cite{midi}
& \Checkmark
& 0.0832 & 0.1073 & 52.38 & 60.43 & 0.2357
& 10.97 & 0.8139 & 0.3867  & 0.5894  \\

SceneGen~\cite{scenegen}
& \Checkmark
& 0.0678 & 0.0616 & 58.33 & {\bf71.89} & 0.5785
& 10.36 & 0.7708 & 0.3719  &0.6050  \\

SAM3D~\cite{sam3d}
& \Checkmark
& 0.0776 & 0.0593 & 71.54 & 60.58 & 0.5813
& 11.59 & 0.8183 & 0.3391  & 0.6096 \\

3D-Fixer~\cite{3D-Fixer}
& \Checkmark
& 0.0692 & 0.0844 & 55.67 & 62.03 & 0.4476
& 10.48 & 0.7877 & 0.3846  & {\bf 0.6270} \\

\midrule
{\bf F3V (ours)}
& \Checkmark
& {\bf 0.0258} & {\bf 0.0579} & {\bf 75.16} & { 69.56} & {\bf 0.5904}
& {\bf 11.63}  & {\bf 0.8262} & {\bf 0.3320}  & 0.6121 \\
\bottomrule
\end{tabular}%
}
\vspace{-8pt}
\end{table*}

\noindent \textbf{Baselines.}
We compare our model with representative approaches for single-image and scene-level 3D generation, including PartCrafter~\cite{Partcrafter}, Gen3DSR~\cite{gen3dsr}, MIDI~\cite{midi}, SceneGen~\cite{scenegen}, SAM3D~\cite{sam3d}, and 3D-Fixer~\cite{3D-Fixer}. For approaches that support mask-guided generation, the corresponding target-object masks are provided as input. For methods without explicit instance-level control, such as PartCrafter, we follow their original inference protocols by supplying cropped object images or object number information when required. Since several baselines do not provide complete texture generation or rendering implementations, visual comparisons are performed only on methods with reproducible rendering results under the same evaluation conditions. The purpose of the comparison is not to show superiority of our texture synthesis component over dedicated 3D generators, but to validate whether decoupled layout reasoning improves object positioning, scale estimation, and spatial relationship modeling.

\noindent \textbf{Benchmarks.}
We conduct geometric and visual evaluations on the 3D-FUTURE test set~\cite{3d-future}. Each test sample contains a photorealistic scene image, target objects with segmentation annotations, and corresponding 3D ground-truth information for quantitative evaluation. The dataset covers diverse indoor furniture categories and various object arrangements, making it suitable for assessing single-image scene reconstruction and object-level layout estimation. In addition, we construct a dedicated validation benchmark for physical attribute evaluation. Specifically, physically controllable assets~\cite{Physx-3d} are retrieved according to the layouts in the 3D-FUTURE test set, enabling evaluation of the accuracy of scene-level physical properties predicted by our framework.

\begin{figure}[t]
 \includegraphics[width=\linewidth]{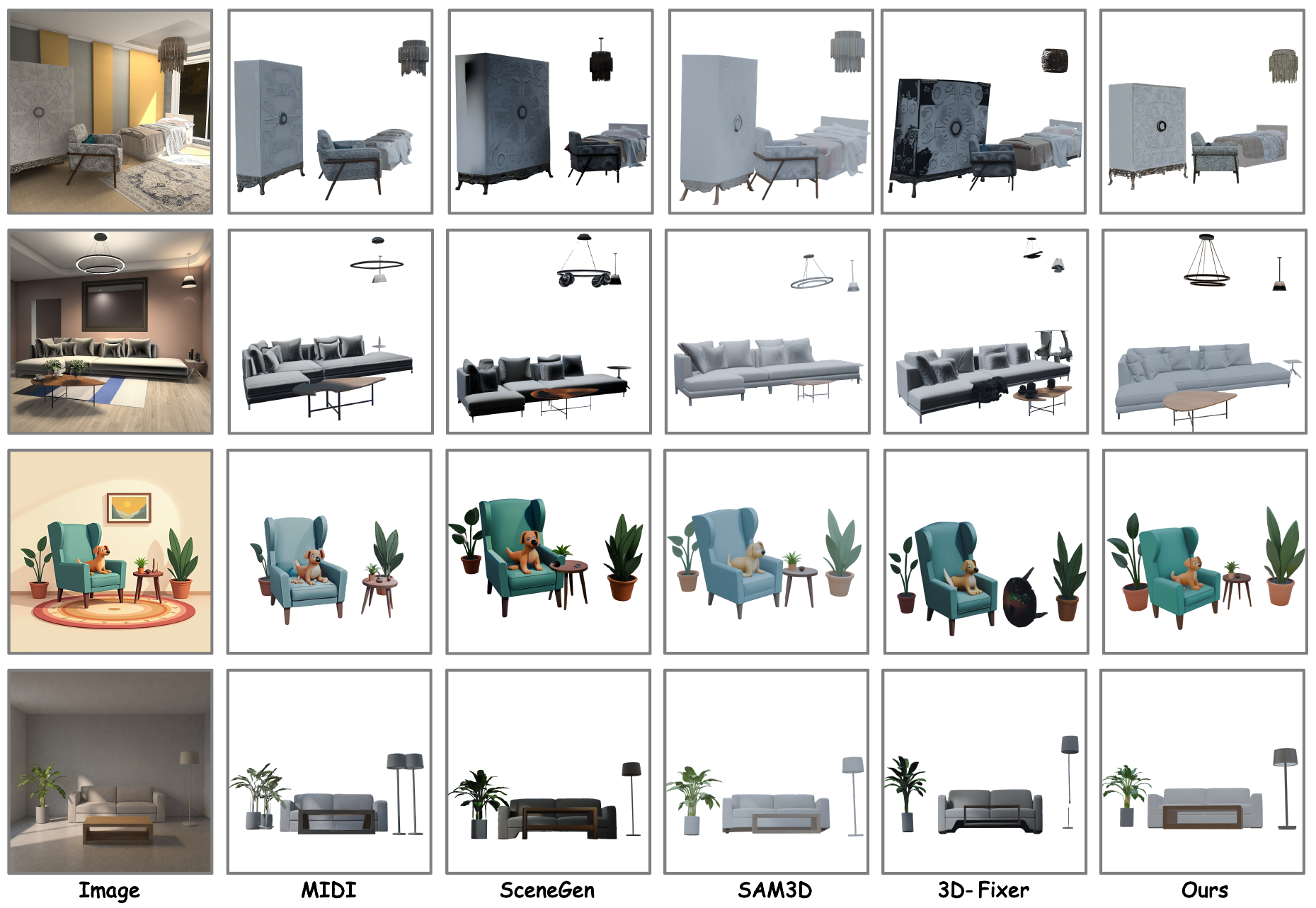}
\caption{Qualitative comparison of scene reconstruction results under diverse environments.
The examples include in-domain and out-of-domain scenes with different object arrangements and visual appearances. Compared with existing approaches, F3V better preserves object positions, scales, and spatial relationships, demonstrating stronger geometric reasoning and layout consistency.
}
\label{fig4}
\end{figure}

\begin{figure}[t]
 \includegraphics[width=0.6\linewidth]{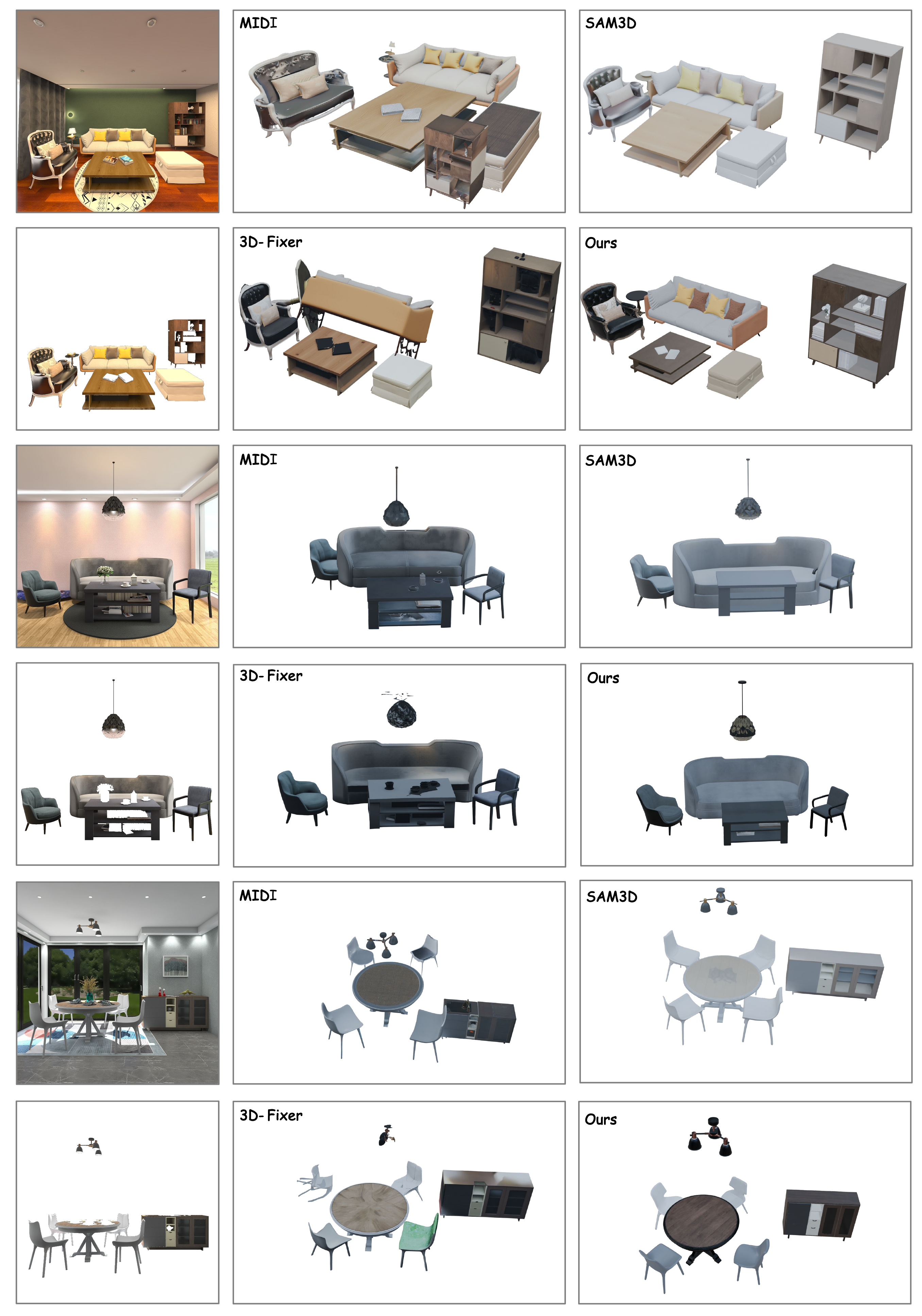}
\caption{Additional qualitative examples. The results demonstrate the robustness of Fysiverse-3D-Vision under diverse scene configurations and object arrangements.}
\label{fig6}
\end{figure}

\subsection{Quantitative Results}

Table~\ref{tab1} summarizes the geometric and visual evaluation results on the 3D-FUTURE test set. F3V achieves consistently strong performance across both reconstruction accuracy and rendering quality, demonstrating the effectiveness of unified spatial modeling for image-based 3D scene reconstruction.

For geometric evaluation, our method obtains the lowest CD-S and CD-O values among all compared approaches, indicating more accurate recovery of scene structures and object-level shapes. The improvements in F-Score-S and IoU-B further verify that the predicted layouts better preserve global spatial organization and object occupancy compared with existing generation-based methods. In particular, the superior IoU-B performance reflects the advantage of decoupling layout reasoning from asset synthesis, allowing object positions and scales to be inferred through explicit geometric understanding rather than relying only on generation priors.

For the visual evaluation, F3V achieves the best PSNR, SSIM, and LPIPS results, showing that improved geometric alignment also benefits image-level appearance consistency. Although 3D-Fixer obtains a slightly higher CLIP-S score, our method maintains competitive semantic similarity while achieving stronger geometric reconstruction. These results demonstrate that geometry-aware spatial reasoning provides a better balance between semantic consistency and physical scene fidelity.

\begin{table}[t]
\caption{
\label{tab2}
Ablation analysis of the proposed multi-stage training strategy.
Each stage is progressively introduced to examine its contribution to geometric reconstruction and spatial layout prediction.
}
\scriptsize
\centering
\resizebox{0.95\linewidth}{!}{%
\begin{tabular}{@{}ccc|ccccc@{}}
\toprule
\multicolumn{3}{c|}{\bf Components} &
\multicolumn{5}{c}{\bf Metrics} \\
{\bf Step 1} & {\bf Step 2} & {\bf Step 3}
& {\bf CD-S $\downarrow$} &
{\bf CD-O $\downarrow$} &
{\bf F-Score-S $\uparrow$} &
{\bf F-Score-O $\uparrow$} &
{\bf IoU-B $\uparrow$} \\
\midrule
\Checkmark &  & 
& 0.1321 & 0.1117 & 42.40 & 52.07 & 0.3551 \\

 & \Checkmark & 
& 0.2495 & 0.1824& 36.52 & 25.18 & 0.3186 \\

 &  & \Checkmark
& 0.0865 & 0.0773 & 58.75 & 59.30 & 0.3855 \\

\Checkmark & \Checkmark & 
& 0.0796 & 0.0814 & 64.67 & 57.02 & 0.4530 \\

\midrule
\Checkmark & \Checkmark & \Checkmark
& {\bf 0.0258} & {\bf 0.0579} & {\bf 75.16} & {\bf 69.56} & {\bf 0.5904}
 \\
\bottomrule
\end{tabular}%
}
\end{table}

\subsection{Qualitative Results}

Figure~\ref{fig4} presents qualitative comparisons on both in-domain and out-of-domain scenes. Compared with existing approaches, the proposed method produces more reliable object arrangements under complex spatial configurations. Baseline methods are generally capable of generating objects that match the semantic categories in the input image, but they frequently exhibit inaccurate object positions, unrealistic scales, missing instances, and incorrect support relationships. Such errors become particularly apparent in scenes containing multiple interacting objects, where local appearance similarity is insufficient to recover the underlying spatial structure.
In contrast, our approach better preserves object co-occurrence patterns and scene-level organization. As illustrated in bedroom, living-room, cartoon-style, and gray-scale indoor scenarios, the predicted layouts maintain more accurate relative distances and spatial relationships among furniture and surrounding objects. This advantage results from the interaction between object-conditioned representations and global geometric features, enabling the model to reason about object placement beyond visual appearance alone.

Additional examples in Figure~\ref{fig6} further demonstrate the generalization ability of our method across diverse scene structures. Compared with existing approaches, our method maintains more stable object placement and fewer geometric inconsistencies, especially in scenes with complex object interactions. These observations indicate that explicit spatial reasoning is critical for constructing usable 3D environments, where semantic recognition alone cannot guarantee physically plausible scene reconstruction.

\subsection{Ablation Studies}

Table~\ref{tab2} investigates the contribution of each component in the proposed three-stage optimization strategy. The results demonstrate that progressively introducing geometry learning, layout reasoning, and physical refinement is essential for achieving accurate and executable scene reconstruction.
The first stage establishes the shared geometry-language representation, which provides semantic and geometric priors for subsequent layout prediction. When only the first stage is applied, the model already obtains meaningful reconstruction ability, confirming that joint semantic and geometric learning provides a strong foundation for spatial understanding. In contrast, directly optimizing the layout module without the learned shared representation leads to significantly degraded performance, indicating that layout prediction cannot be effectively solved without sufficient geometric and semantic context.

The second stage injects layout supervision while preserving the geometry reconstruction objective. Compared with independent layout optimization, this strategy enables the layout module to exploit the geometry-aware representation learned in the previous stage. The improvements in CD-S, F-Score-S, and IoU-B demonstrate that maintaining geometric constraints prevents the model from overfitting to object placement and improves global scene consistency.

The final stage introduces collision-aware refinement using real-world scene constraints. The consistent gains across all evaluation metrics show that explicit physical regularization further improves object arrangement and spatial plausibility. These results verify that the three stages are complementary: geometry-language pretraining provides general spatial knowledge, layout injection transfers this knowledge to object placement, and collision-aware refinement enhances the physical validity of the reconstructed scenes.

\begin{table}[t]
\footnotesize
\centering
\begin{tabular}{@{}l|ccccc@{}}
\toprule
\multirow{2}{*}{\bf Methods} &
\multicolumn{5}{c}{\bf Physical Attributes} \\

& {\bf Scale $\downarrow$}
& {\bf Mat. $\uparrow$}
& {\bf Aff. $\uparrow$}
& {\bf Kin. $\uparrow$}
& {\bf Desc. $\uparrow$} \\
\midrule

MIDI~\cite{midi}
& --- & --- & --- & --- & --- \\
MIDI + PhysPre
& 12.54 & 7.17 & 10.03 & 0.27 & 7.05 \\

\midrule
\textbf{F3V (ours)}
& \bf 8.09 & \bf 14.51 & \bf 11.90 & \bf 0.32 & \bf 10.94\\
\bottomrule
\end{tabular}
\caption{Evaluation of physical attribute understanding on executable 3D assets.
Scale denotes absolute-scale error, while the remaining metrics measure the prediction accuracy of material, affordance, kinematic properties, and textual descriptions.
}
\label{table3}
\end{table}

\subsection{Physical-Aware Representation and Executable Reconstruction}

Following the evaluation setting of PhysX-3D, Table~\ref{table3} provides a comparison of different approaches on physical attribute prediction and executable asset understanding. Compared with MIDI and its variants enhanced with physical priors, our method achieves consistent improvements across multiple aspects, including absolute scale estimation, material recognition, affordance prediction, kinematic modeling, and textual description generation. These results demonstrate that the shared vision-language-geometry representation captures not only spatial configurations but also object-level functional knowledge and physical characteristics, which are essential for constructing interactive and executable 3D environments.

\begin{figure}[t]
 \includegraphics[width=1\linewidth]{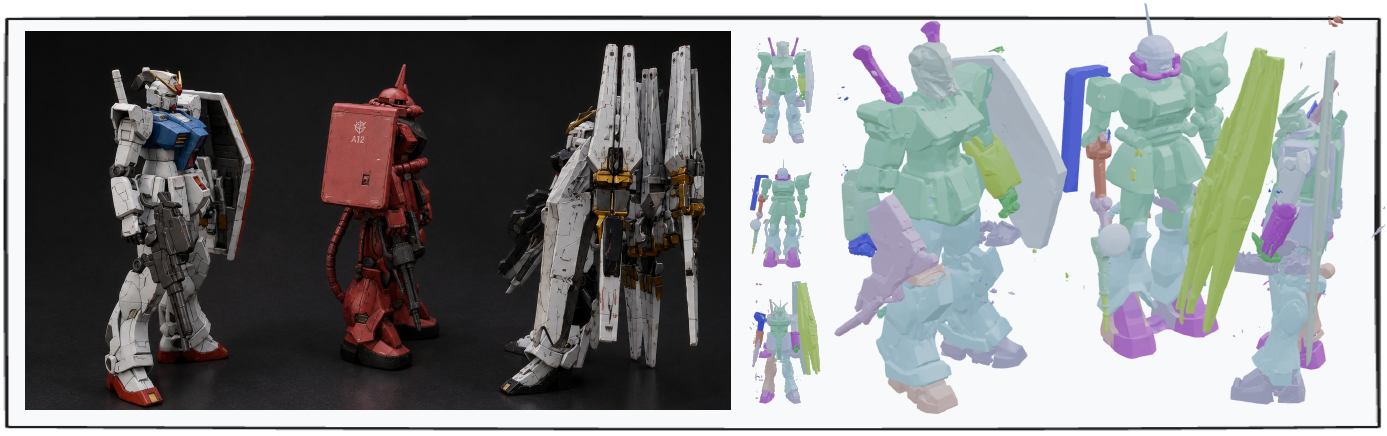}
\caption{Extension to part-level executable scene reconstruction.
By integrating part-aware asset generation with the predicted spatial layout, F3V enables fine-grained reconstruction beyond object-level placement.
}
\label{fig5}
\end{figure}

Beyond object-level physical understanding, F3V can be further extended to fine-grained interactive scene reconstruction with part-level annotations. As illustrated in Figure~\ref{fig5}, combining our spatial layout prediction with part-aware asset generation methods such as PartCrafter enables scene reconstruction with detailed structural decomposition. This extension improves the granularity of executable scene representation from complete objects to individual parts. In contrast, existing part-level generation approaches mainly focus on isolated object synthesis and may lose fine-grained structures when applied to scene-level reconstruction. The ability to preserve both scene context and part-level details highlights the advantage of our unified spatial representation.

\section{Conclusion}

We present Fysiverse-3D-Vision, a unified vision-language-geometry framework for executable 3D scene reconstruction from a single image. Different from existing approaches that tightly couple spatial layout estimation with object generation, our framework decouples layout reasoning from asset synthesis and learns object placement from shared semantic and geometric representations. By integrating textual understanding, visual semantics, and geometric structures within a unified architecture, our model is able to jointly capture scene context, metric geometry, and object-level spatial relationships.
The proposed multi-stage training strategy progressively builds geometry-language representations, injects layout reasoning while preserving reconstruction capability, and improves physical consistency through collision-aware refinement. Extensive experiments demonstrate that the method achieves superior performance in geometric reconstruction, spatial layout estimation, visual quality, and physical attribute understanding. Additional evaluations on executable assets and part-level reconstruction further verify the flexibility of the learned spatial representation for interactive scene construction.


\bibliographystyle{plainnat}
\bibliography{main}


\end{document}